\documentclass[sigconf]{acmart}

\AtBeginDocument{%
  \providecommand\BibTeX{{%
    \normalfont B\kern-0.5em{\scshape i\kern-0.25em b}\kern-0.8em\TeX}}}

\copyrightyear{2023}
\acmYear{2023}
\setcopyright{acmlicensed}\acmConference[MM '23]{Proceedings of the 31st ACM International Conference on Multimedia}{October 29-November 3, 2023}{Ottawa, ON, Canada}
\acmBooktitle{Proceedings of the 31st ACM International Conference on Multimedia (MM '23), October 29-November 3, 2023, Ottawa, ON, Canada}
\acmPrice{15.00}
\acmDOI{10.1145/3581783.3612008}
\acmISBN{979-8-4007-0108-5/23/10}

\acmSubmissionID{1382}

\usepackage{multirow}
\usepackage{algorithm}
\usepackage{algorithmicx}
\usepackage{algpseudocode}
\usepackage{amsmath}
\usepackage{mathtools}
\usepackage{url}
\usepackage{soul}
\usepackage{balance}

\usepackage{graphicx}
\usepackage{amssymb}
\usepackage{booktabs}
\usepackage[percent]{overpic}

\usepackage{xcolor}
\usepackage{makecell}

\usepackage[capitalize]{cleveref}
\crefname{section}{Sec.}{Secs.}
\Crefname{section}{Section}{Sections}
\Crefname{table}{Table}{Tables}
\crefname{table}{Tab.}{Tabs.}

\usepackage{subcaption}

\usepackage{enumitem}

\UseRawInputEncoding

\begin{document}

\title[CoColor]{Cooperative Colorization: Exploring Latent Cross-Domain Priors for NIR Image Spectrum Translation}



\author{Xingxing Yang}
\affiliation{%
  \institution{Department of Computer Science, Hong Kong Baptist University}
  \streetaddress{Kowloon Tong}
  \country{Hong Kong SAR, China}
}
\email{csxxyang@comp.hkbu.edu.hk}

\author{Jie Chen}
\authornote{Corresponding author: Jie Chen}
\affiliation{%
  \institution{Department of Computer Science, Hong Kong Baptist University}
  \streetaddress{Kowloon Tong}
  \country{Hong Kong SAR, China}
}
\email{chenjie@comp.hkbu.edu.hk}

\author{Zaifeng Yang}
\affiliation{%
  \institution{Institute of High Performance Computing, A*STAR}
  \streetaddress{Singapore}
  \country{Singapore}
}
\email{yang\_zaifeng@ihpc.a-star.edu.sg}

\renewcommand{\shortauthors}{Xingxing Yang, Jie Chen, \& Zaifeng Yang}
\begin{abstract}
Near-infrared (NIR) image spectrum translation is a challenging problem with many promising applications. Existing methods struggle with the mapping ambiguity between the NIR and the RGB domains, and generalize poorly due to the limitations of models' learning capabilities and the unavailability of sufficient NIR-RGB image pairs for training. 
To address these challenges, we propose a cooperative learning paradigm that colorizes NIR images in parallel with another proxy grayscale colorization task by exploring latent cross-domain priors (i.e., latent spectrum context priors and task domain priors), dubbed CoColor. The complementary statistical and semantic spectrum information from these two task domains -- in the forms of pre-trained colorization networks -- are brought in as task domain priors. A bilateral domain translation module is subsequently designed, in which intermittent NIR images are generated from grayscale and colorized in parallel with authentic NIR images; and vice versa for the grayscale images. These intermittent transformations act as latent spectrum context priors for efficient domain knowledge exchange. We progressively fine-tune and fuse these modules with a series of pixel-level and feature-level consistency constraints.
Experiments show that our proposed cooperative learning framework produces satisfactory spectrum translation outputs with diverse colors and rich textures, and outperforms state-of-the-art counterparts by \textbf{3.95dB} and \textbf{4.66dB} in terms of PNSR for the NIR and grayscale colorization tasks, respectively.
\end{abstract}



\maketitle

\begin{figure}[H]
    \centering
  { 
      \includegraphics[width=1\linewidth]{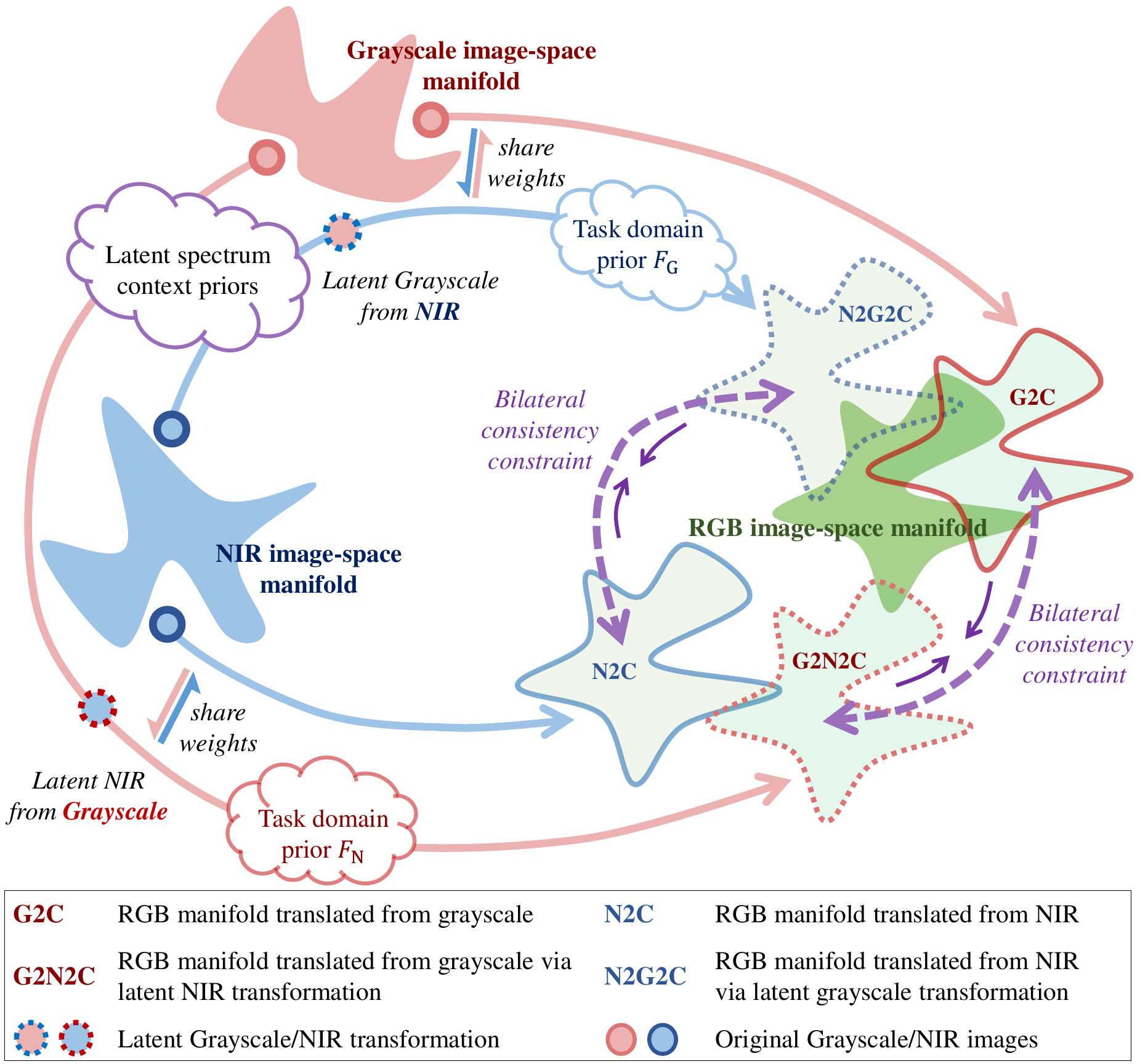}}
      \caption{Description of the cooperative learning paradigm. For a NIR source image, the bilateral domain translation module, the grayscale image colorization module, and the NIR image colorization module are pre-trained to generate two approximated manifolds (N2C and N2G2C). Subsequently, the whole model is fine-tuned with the bilateral consistency constraint, which keeps bringing the two approximated manifolds closer together toward the target RGB image-space manifold.}
      \label{manifold} 
\end{figure}

\section{Introduction}\label{sec:introduction}

Near-infrared (NIR) imaging focuses on capturing the electromagnetic spectrum details of the scene, covering the wavelength range from 780nm to 1000nm.  
With such distinctive spectrum coverage, NIR images provide unique values in capturing subtle visual details, especially under low-light and foggy atmospheric conditions, making NIR imaging popular in applications such as nighttime video surveillance \cite{balci2019front}, object detection \cite{elihos2018comparison}, and remote sensing \cite{protopapadakis2021stacked}. 
However, the unique spectrum response of NIR images is not similar to human vision and computer vision systems, both of which are accustomed to/trained over scene reflectance under visible light. To visualize the NIR images more naturally and intuitively, NIR-to-RGB spectral domain translation (i.e., NIR image colorization) has become a valuable research topic.



\par

Significant advancement has been achieved in recent years on the tasks of grayscale image colorization \cite{yatziv2006fast}, \cite{li2017example}, \cite{fang2019superpixel}, \cite{su2020instance}, \cite{kumar2021colorization};
however, the progress of NIR image colorization falls behind. Results from recently reported studies \cite{liang2021improved}, \cite{suarez2017infrared}, \cite{dong2018infrared}, \cite{9301788} are unsatisfactory in terms of color realism and diversity, and content/texture fidelity. 
We argue that NIR image colorization is a much more challenging problem than grayscale image colorization, and the reasons are analyzed as follows:
\par
(\textbf{\romannumeral1}) \textbf{Mapping Ambiguity.} 
Grayscale images can be transformed from the RGB domain deterministically with a projection matrix, and only chrominance values need to be estimated as a single-to-many inverse problem for RGB colorization.
In contrast, there is no explicit relationship between NIR and RGB domains. Since the NIR spectral band ($780-1000nm$) is distinct from RGB (blue: $450-485nm$, green: $500-565nm$, and red: $625-750nm$), the mapping relationship is much more nonlinear compared with grayscale image colorization.
To translate RGB images from the NIR domain, both the chrominance and luminance values need to be estimated, making it a much more challenging \textit{many-to-many} inverse mapping problem. 


\par
(\textbf{\romannumeral2}) \textbf{Poor Generalization.} Unlike almost unlimited grayscale-RGB image pairs, few datasets provide NIR-RGB image pairs, which require professional capture equipment and dedicated algorithms to align the pixels. 
This also limits the number of NIR-RGB image pairs for each dataset. For instance, ImageNet \cite{yang2019fairer} provides millions of grayscale-RGB image pairs, whilst NIR-RGB image datasets are usually much smaller. For example, the VCIP dataset $\footnote{Website of this challenge: \url{http://www.vcip2020.org/grand_challenge.htm}}$ and the EPFL dataset \cite{brown2011multi} both provide only hundreds of image pairs. 
The lack of training data usually leads to poorly generalized and over-fitted models  \cite{hawkins2004problem}.
\par
\par
A straightforward question arises: \textit{can the NIR image spectrum translation task learn from the grayscale image colorization task?} 
To address this issue, we propose to explore the latent cross-domain priors, i.e., \textit{task domain priors} and \textit{latent spectrum context priors}. As shown in Fig.~\ref{manifold}, we propose a cooperative learning paradigm that first brings in pre-trained NIR and grayscale colorization networks as \textit{task domain priors}. In order to efficiently fuse the statistical and semantic knowledge from both task domains, we further propose a bilateral domain translation module, in which intermittent NIR images are generated from grayscale and colorized in parallel with authentic NIR images; and vice versa for the grayscale images. These intermittent transformations act as \textit{latent spectrum context priors} for efficient domain knowledge exchange. Finally, we progressively fine-tune and fuse these modules with a series of pixel-level and feature-level consistency constraints.

\begin{figure}[t]
    \centering
  { 
      \includegraphics[width=1\linewidth]{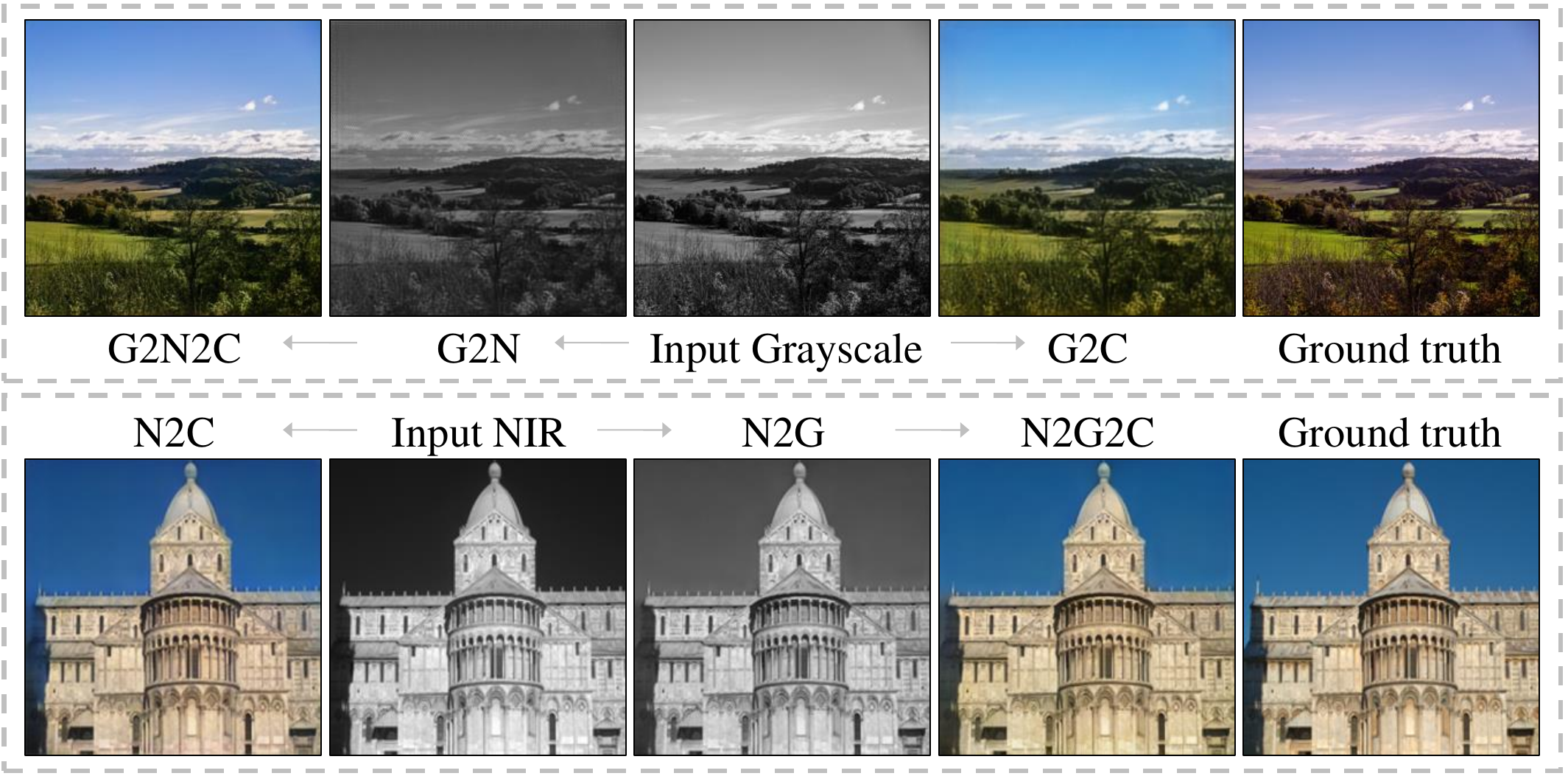}}
      \caption{Colorization results of our method on both grayscale and NIR image colorization tasks. \textbf{G2N2C}: \textit{gray-to-NIR-to-RGB}, \textbf{G2N}: \textit{gray-to-NIR}, \textbf{G2C}: \textit{gray-to-RGB}; \textbf{N2C}: \textit{NIR-to-RGB}, \textbf{N2G}: \textit{NIR-to-gray}, \textbf{N2G2C}: \textit{NIR-to-gray-to-RGB}.}
      \label{illustration} 
\end{figure}

Notably, the two colorization tasks complement each other and mutually improve: while one of our task domain priors -- the grayscale colorization prior brings in stable semantic knowledge from large datasets, the NIR counterpart complements texture details and spectral mapping diversity. This effectively resolves the aforementioned challenges for NIR-RGB translation: mapping ambiguity and poor generalization . 
An illustration of the colorization outputs is shown in Fig. \ref{illustration} with realistic, diverse colors and well-preserved textures. 
The main contributions of this work can be generalized as follows:
\begin{itemize}
\item We propose a cooperative learning paradigm that colorizes NIR images via parallel and cooperative learning with another grayscale colorization module, which acts as the task domain prior to tackle the challenges of the mapping ambiguity between the NIR and the RGB domains, and poor generalization due to the unavailability of sufficient NIR and RGB training image pairs.
\item We propose a bilateral domain translation module to generate latent spectrum translations, which serve as the latent spectrum context priors to progressively fuse the spectrum knowledge between the two task domains.
\item We propose a progressive and cooperative training strategy to fine-tune the three pre-trained modules cooperatively to fuse the statistical and semantic knowledge from both task domain priors and latent spectrum context priors (these two priors constitute our latent cross-domain priors) efficiently and achieve a new state-of-the-art performance.

\end{itemize}

\section{Related Works}

\begin{figure*}[t]
    \centering
  { 
      \includegraphics[width=0.96\linewidth]{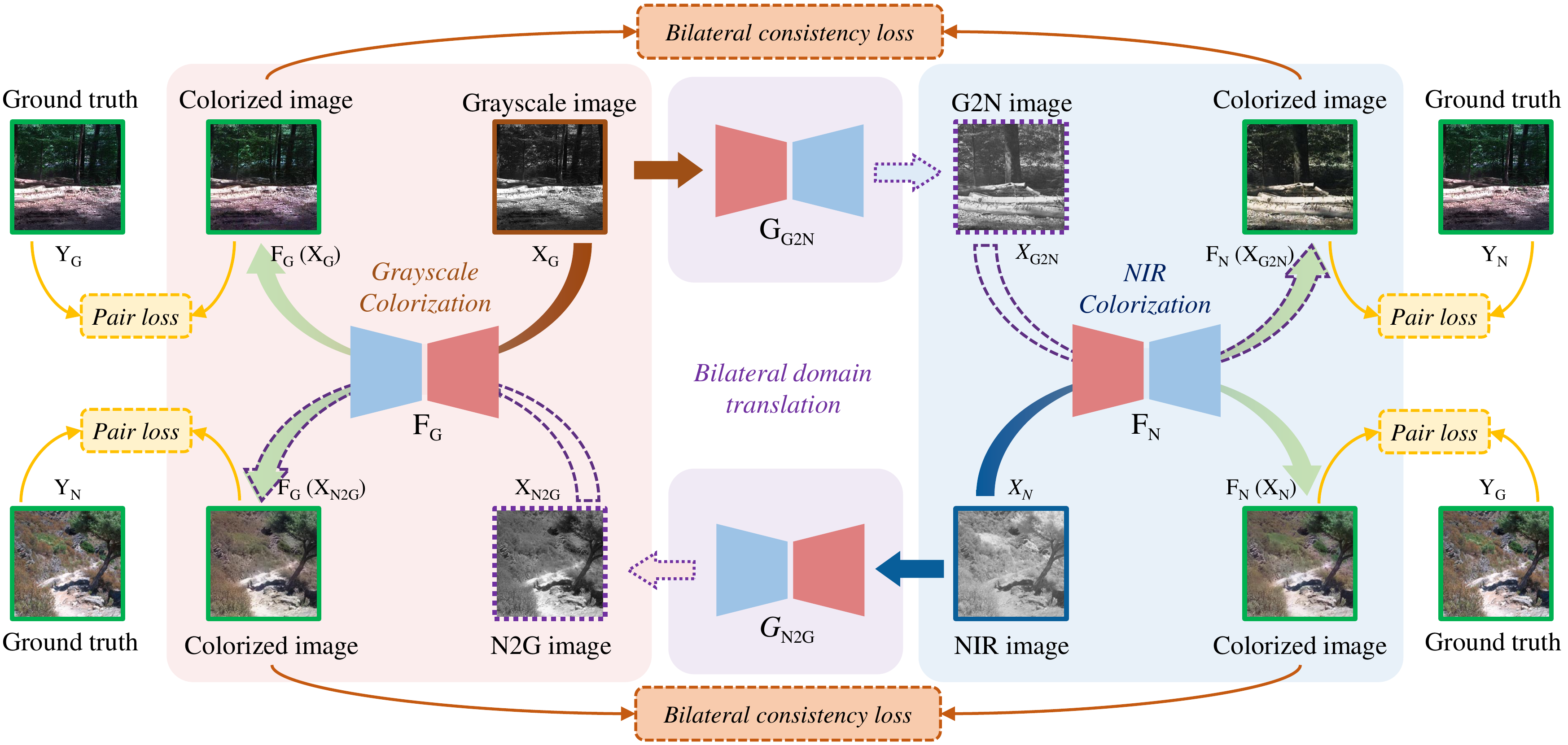}}
      \caption{Framework illustration.  It includes two parts: a bilateral domain translation module that translates grayscale images and NIR images into each other to generate latent spectrum translations (N2G image and G2N image), which act as the latent spectrum context priors; and two parallel colorization modules (NIR and grayscale image colorization) that perform colorization on each domain parallelly and serve as the task domain priors ($F_G(X_{N2G})$ and $F_N(X_{G2N})$) to each other.}
      \label{architure} 
\end{figure*}

\textbf{NIR Spectrum Translation.} 
Existing methods mainly focused on GAN-based \cite{goodfellow2014generative} frameworks to align generated results with RGB ground truths at a pixel level~\cite{10.1007/978-3-319-61578-3_16}, \cite{liang2021improved}, \cite{suarez2017infrared}, \cite{dong2018infrared}. In general, these methods can be divided into three categories: supervised, unsupervised, and semi-supervised paradigms. 
For supervised methods, \cite{10.1007/978-3-319-61578-3_16} utilized DCGAN to implement NIR image colorization and generated colorful results. However, the network is not able to learn high-level features in both the NIR domain and the RGB domain, resulting in some blurs and color distortions in generated results. 
Moreover, these methods are all trained on paired NIR and RGB datasets (only several hundred image pairs are available), where small-size training sets limit the performance of learning-based methods. 
To this end, \cite{9025662} leveraged CycleGAN \cite{Conventional_CycleGAN} based on unpaired data to align gaps of both NIR and RGB domains at a pixel level. \cite{9301791} further proposed an alternatively trained CycleGAN \cite{Conventional_CycleGAN} to utilize both paired and unpaired data, and introduced cross-scale skip connections into the decoder of the UNet-based generator, which could generate more vivid colorized results, but still failed in predicting results which were ``more loyal'' to ground truths and lost some texture information.
We argue that either directly translating the NIR into the RGB domain or using CycleGAN \cite{Conventional_CycleGAN} to align the domain gap between NIR and RGB images in pixels is suboptimal because of the intrinsic mapping ambiguity and non-ignorable domain gaps. 

\noindent
\textbf{Image Prior.}
Image priors have been applied in many restoration and translation tasks, which describe the image statistics such as semantics \cite{nguyen2019semantic} and texture \cite{pickup2003sampled}. Traditionally handcrafted image priors, \textit{e.g.}, dark channel prior (DCP) \cite{he2010single}, \cite{pan2016blind} and total variation (TV) criterion \cite{zhang2015variational}, are used in many ill-posed inverse problems since they provide a systematic and comprehensive framework to estimate model parameters. 
Recently, learning-based image priors \cite{Yang_2021_CVPR}, \cite{ulyanov2018deep}, \cite{pan2021exploiting} have attracted extensive attention, showing that deep neural networks capture image statistics implicitly with more non-linear estimation than handcrafted image priors. Specifically, for image colorization tasks, reference-based methods \cite{lu2020gray2colornet}, \cite{welsh2002transferring}, \cite{ironi2005colorization} learn the semantic of reference images and then transfer it into target images. Yanze Wu \textit{et al.} \cite{Wu_2021_ICCV} proposed to extract diverse color priors of the target image itself so that the priors could be encapsulated in a pre-trained GAN and then fused them into a colorization module by feature modulations. 
However, these priors extraction methods do not apply to NIR images colorization since they all belong to the same spectrum domain, while we aim to exploit the latent cross-domain priors between grayscale and NIR images, where both visual and spectral domain discrepancies exist.

\section{Proposed Method}

\textbf{Motivation.}
As mentioned earlier, the mapping from NIR to RGB domain is not bijective. Consequently, direct regression often leads to monotonous outputs (See the results produced by existing methods in Figure. \ref{visual comparison}). A common approach to dealing with such a challenge is to incorporate a latent variable $Z$. The posterior probability for the RGB prediction $Y$ could be modeled as a conditional Variational Autoencoder \cite{kingma2013auto}, \cite{10.1145/3306346.3323020} given the input NIR image $X_{\text{N}}$ as:
\begin{footnotesize}
\begin{equation}
p\left(Y | X_{\text{N}}\right)=\int p\left(Y | Z, X_{\text{N}}\right) \cdot p\left(Z | X_{\text{N}}\right) dZ.
\end{equation}
\end{footnotesize}
There are multiple choices for the latent variable $Z$, which functions to disambiguate the spectrum mapping. For instance, a semantic segmentation map could be predicted and employed as $Z$ to bring in additional semantic information for the colorization task.
In contrast to such a straightforward choice, we aim to take full advantage of the latent spectrum context priors from grayscale-RGB data pairs, which are abundantly available. In specific, we predict a latent NIR image translated from grayscale: $X_{\text{N2G}}$, as the latent variable $Z$:
\begin{footnotesize}
\begin{align}
\label{domain translation description}
p\left(Y | X_{\text{N}}\right) &=\int p\left(Y | X_{\text{N2G}}, X_{\text{N}}\right) \cdot p\left(X_{\text{N2G}} | X_{\text{N}}\right) dX_{\text{N2G}}.
\end{align}
\end{footnotesize}
Our consideration is threefold:
\par
\textbf{(i) High-resolution Latent Textures are introduced.} Compared with a compressed representation (\textit{e.g.}, latent vectors used in \cite{10.1145/3306346.3323020}, the latent spectrum translation $X_{\text{N2G}}$ has a higher resolution, making it possible to capture more accurate local texture details. Thus, the texture information from both NIR and grayscale domains can be fully used to infer the local semantics and appearance changes in the scene representations.
\par
\textbf{(ii) Simple Posterior $p(X_{\text{N2G}}| X_{\text{N}})$ and $p(Y | X_{\text{N2G}})$ are available.}
During training, the ground truth of the latent spectrum translation $X_{\text{N2G}}$ for each training frame can be obtained by transforming the RGB image (from the RGB-NIR image pairs) to grayscale:~$X_{\text{G}}$. Consequently, learning of the task $p(Y|X_{\text{N}})$ can be split into two parts: the \textit{latent spectrum context prior} $p(X_{\text{N2G}} | X_{\text{N}})$ and the proxy-task colorization $(Y | X_{\text{N2G}})$ (defined as the \textit{task domain prior}). Similarly, learning of the proxy-task $p(Y|X_\text{{G}})$ can also be split into two parts: the other \textit{latent spectrum context prior} $p(X_{\text{G2N}}|X_{\text{G}})$ and the other \textit{task domain prior} $p(Y | X_\text{{G2N}})$.
\par
\textbf{(iii) Information Fusion benefits all tasks.}
With these latent parameters introduced, knowledge learned from each task domain can be complementary to others. Thus, efficient information fusion is crucial. We propose a progressive and iterative training process that encourages efficient information fusion and exchange between the learning of domain priors and context priors, which resolves ambiguities for each individual learning track.

\noindent
\textbf{Method Overview.} As shown in Fig.~\ref{architure}, the proposed CoColor framework comprises two colorization modules ($F_\text{N}$ and $F_\text{G}$) and a bilateral domain translation module ($G_{\text{G2N}}$ and $G_{\text{N2G}}$). Specifically, the bilateral domain translation module generates the latent grayscale translation from NIR images $X_{\text{N2G}}=G_{\text{N2G}}(X_{\text{N}})$ and the latent NIR translation from grayscale images $X_{\text{G2N}} =G_{\text{G2N}}(X_{\text{G}})$, constituting the spectrum context priors.
With the latent spectrum translations, the NIR image colorization module $F_\text{N}$ translates both the real and the latent NIR images (i.e., $X_\text{N}$ and $X_{\text{G2N}}$) to the RGB domain; while the grayscale image colorization module $F_\text{G}$ translates both the real and the latent grayscale images (i.e., $X_\text{G}$ and $X_{\text{N2G}}$) to the RGB domain, constituting the task domain priors. 
By incorporating these latent cross-domain priors, the statistical and semantic knowledge from both NIR and grayscale domains are efficiently fused with a series of pixel-/feature-level consistency constraints.




\subsection{Colorization Modules: NIR and Grayscale}
As introduced, the colorization modules $F_\text{G}$ and $F_\text{N}$ perform grayscale and NIR image colorization, respectively. 
By obtaining the latent spectrum translations from the bilateral domain translation module (to be introduced in Sec. \ref{sec_bilateral}), $F_\text{G}$ colorizes both real $X_\text{G}$ and the latent $X_\text{N2G}$ grayscale images translated from the NIR domain, while $F_\text{N}$  colorizes both real $X_\text{N}$ and the latent $X_{\text{G2N}}$  NIR images translated from the grayscale domain.
The generators for both the NIR and grayscale colorization modules share similar encoder-decoder architectures based on the `U-Net256’ structure specified in \cite{9301791}. More structure details are provided in the \textbf{\textit{Supplementary Material}}. 


\subsection{Bilateral Domain Translation Module} \label{sec_bilateral}
As illustrated in Fig.~\ref{architure}, the bilateral domain translation module comprises a \textit{gray-to-NIR} translator $G_{\text{G2N}}$ and a \textit{NIR-to-gray} translator $G_{\text{N2G}}$, both working to establish the \textit{latent spectral connections} between the NIR and grayscale domains. 
In specific, $G_{\text{G2N}}$ takes grayscale images as input and generates the latent NIR translation $X_{\text{G2N}}$; $G_{\text{N2G}}$ takes NIR images as input and generates the latent grayscale translation $X_{\text{N2G}}$.

The purpose for proposing the bilateral domain translation module is threefold:
i) to generate latent NIR translations from grayscale images and help with the problem of not having enough NIR-RGB data pairs for model training; ii) to relieve the challenge of spectrum mapping ambiguity of the NIR colorization task by co-learning with the grayscale colorization task, which is comparatively more stable in semantic interpretation; and iii) to introduce diversity to the grayscale colorization task by introducing source variations from the latent NIR translations.

\begin{figure}[t]
    \centering
  { 
      \includegraphics[width=1\linewidth]{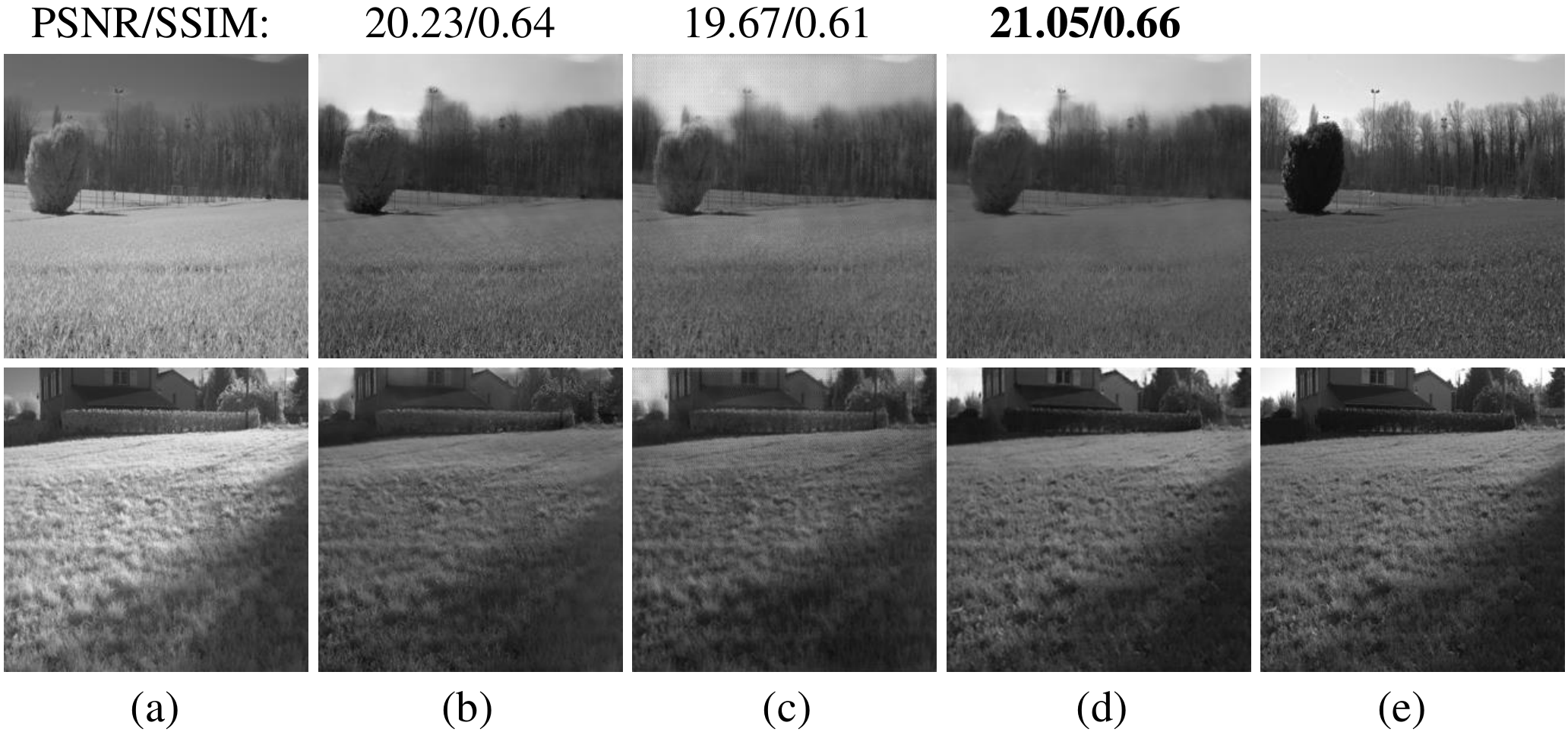}}
      \caption{Intermediate latent spectrum translation results of the bilateral domain translation module. (a) are NIR input images, (b) are pre-trained results, (c) are results trained from scratch, (d) are cooperatively trained results, and (e) are grayscale ground truths.}
      \label{evaluation_DA} 
\end{figure}

\textbf{Remark.} The outputs of the bilateral domain translation modules are not directly used to produce final results but as \textit{intermediate guidance} for the two colorization modules. 
However, these latent translations still achieve $20.23$dB PSNR and $0.64$ SSIM in the testing dataset of the EPFL dataset, and the visualization can be found in Fig. \ref{evaluation_DA}(b). These demonstrate that our bilateral domain translation module effectively aligns the domain discrepancy of NIR and grayscale images. 
\par

\subsection{Learning Objectives} \label{loss}
To explore the latent cross-domain priors (\textit{i.e.}, latent spectrum context priors and task domain priors) and the spectrum translation functions,
we employ the following loss terms as the learning objectives. 

\noindent
\textbf{Colorization Losses.}
With the outputs from the bilateral domain translation module, we need to colorize both the original ($X_\text{N}$ and $X_\text{G}$) and the latent translated images ($X_{\text{G2N}}$ and $X_{\text{N2G}}$). 
Hence, we employ a \textbf{pair loss} to supervise both colorization processes at a pixel level:
\begin{footnotesize}
\begin{align}
\label{pair loss}
\nonumber L_{\text{pair}} & =\left\|Y_\text{G} - F_\text{G}(X_\text{G}) \right\|_{1} + \lambda_{3}\left\|Y_\text{G} - F_\text{G}(X_{\text{N2G}}) \right\|_{1} \\ & + \left\|Y_\text{N} - F_\text{N}(X_\text{N}) \right\|_{1} + \lambda_{4}\left\|Y_\text{N} - F_\text{N}(X_{\text{G2N}}) \right\|_{1},
\end{align}
\end{footnotesize} 
where $Y_\text{N}$ and $Y_\text{G}$ denote ground truths of grayscale and NIR images, respectively. 

To ensure the colorization consistency between identical input images through different modules, \textit{e.g.}, between direct NIR colorization: $F_n(X_\text{N})$, and indirect colorization via latent grayscale translation: $F_{\text{G}}\left(G_{\text{N2G}}\left(X_{\text{N}}\right)\right)$ (and vice versa for $X_\text{G}$ inputs), we introduce the \textbf{bilateral consistency loss}: 
\begin{footnotesize}
\begin{align}
\label{consistency loss}
\nonumber L_{\text{blt}}&=L_{\text{mix}}(F_{\text{N}}\left(X_{\text{N}}\right), F_{\text{G}}\left(G_{\text{N2G}}\left(X_{\text{N}}\right)\right))\\
&+ L_{\text{mix}}( F_{\text{G}}\left(X_{\text{G}}\right), F_{\text{N}}\left(G_{\text{G2N}}\left(X_{\text{G}}\right)\right)),
\end{align}
\end{footnotesize}
where consistency is encouraged at both pixel level (via L1 norm) and perceptual level (via multi-scale SSIM similarity) through the $L_{\text{mix}}$ loss \cite{7797130}.

\noindent
\textbf{Domain Translation Losses.}
For the \textit{gray-to-NIR} domain translation process, we expect the latent NIR translation from grayscale: $X_{\text{G2N}}$, to be indistinguishable from the real NIR images: $X_\text{N}$.
Consequently, we employ discriminators at both pixel-level: $D_{\text{N}}^{\text{img}}$, and feature-level: $D_{\text{N}}^{\text{feat}}$, using the PatchGAN \cite{isola2017image} architecture. 
The \textbf{adversarial losses} are defined as follows.
\begin{footnotesize}
\begin{align}
\label{GAN_pixel_N}
\nonumber &L_{\text{GAN}}^{\text{img}}\left(X_{\text{N}},X_{\text{G}};D_{\text{N}}^{\text{img}},G_{\text{G2N}}\right) = \mathbb{E}_{\text{x}_{\text{g}} \sim X_{\text{G}}}\left[D_{\text{N}}^{\text{img}}\left(G_{\text{G2N}}(x_{\text{g}})\right)\right] \\  & \quad \quad +\mathbb{E}_{\text{x}_{\text{n}} \sim X_{\text{N}}}\left[D_{\text{N}}^{\text{img}}\left(x_{\text{n}}\right)-1\right],
\end{align}
\begin{align}
\label{GAN_feature_N}
\nonumber &L_{\text{GAN}}^{\text{feat}}\left(\!X_{\text{N}},X_{\text{G}};D_{\text{N}}^{\text{feat}},G_{\text{G2N}},F_{\text{N}}\!\right) \!= \! \mathbb{E}_{\text{x}_{\text{g}} \sim X_{\text{G}}}\!\left[D_{\text{N}}^{\text{feat}}\!\left(F_{\text{N}}(G_{\text{G2N}}(x_{\text{g}}))\right)\!\right] \\ 
& \quad \quad \quad +\mathbb{E}_{\text{x}_{\text{n}} \sim X_{\text{N}}}\left[D_{\text{N}}^{\text{feat}}\left(F_{\text{N}}(x_{\text{n}})\right)-1\right]. 
\end{align}
\end{footnotesize}

Similarly, the \textit{NIR-to-gray} translation process has the following \textbf{adversarial losses}:
\begin{footnotesize}
\begin{align}
\label{GAN_pixel_G}
\nonumber &L_{\text{GAN}}^{\text{img}}\left(X_{\text{N}},X_{\text{G}},D_{\text{G}}^{\text{img}},G_{\text{N2G}}\right)=\mathbb{E}_{\text{x}_{\text{n}} \sim X_{\text{N}}}\left[D_{\text{G}}^{\text{img}}\left(G_{\text{N2G}}(x_{\text{n}})\right)\right] \\ 
& \quad \quad  +\mathbb{E}_{\text{x}_{\text{g}} \sim X_{\text{G}}}\left[D_{\text{G}}^{\text{img}}\left(x_{\text{g}}\right)-1\right], 
\end{align}
\begin{align}
\label{GAN_feature_G}
\nonumber & L_{\text{GAN}}^{\text{feat}}\!\left(\!X_{\text{N}},X_{\text{G}},D_{\text{G}}^{\text{feat}},G_{\text{N2G}},F_{\text{G}}\!\right)\!\!=\!\mathbb{E}_{\text{x}_{\text{n}} \sim X_{\text{N}}}\!\left[D_{\text{G}}^{\text{feat}}\left(F_{\text{G}}(G_{\text{N2G}}(x_{\text{n}}))\!\right)\right] \\ 
& \quad \quad \quad +\mathbb{E}_{\text{x}_{\text{g}} \sim X_{\text{G}}}\left[D_{\text{G}}^{\text{feat}}\left(F_{\text{G}}(x_{\text{g}})\right)-1\right].
\end{align}
\end{footnotesize}
In addition, a \textbf{cycle-consistency loss} \cite{Conventional_CycleGAN} is used to constrain the content consistency of a NIR image $X_\text{N}$, when it is translated by $G_{\text{N2G}}$ to grayscale, and then by $G_{\text{G2N}}$ back to NIR, and vice versa for $X_\text{G}$:
\begin{footnotesize}
\begin{equation}
\begin{aligned}
\label{domain_cycle}
L_{\text{cyc}}&=\mathbb{E}_{\text{x}_{\text{g}} \sim X_{\text{G}}}\left[\| G_{\text{N2G}}\left(G_{\text{G2N}}(x_{\text{g}}))-x_{\text{g}}\right \|_{1}\right] \\
&+\mathbb{E}_{\text{x}_{\text{n}} \sim X_{\text{N}}}\left[\left\|G_{\text{G2N}}\left(G_{\text{N2G}}\left(x_{\text{n}}\right)\right)-x_{\text{n}}\right\|_{1}\right].
\end{aligned}
\end{equation}
\end{footnotesize}

Since images are being translated among multiple domains and along multiple directions, we employ the \textbf{identity loss} \cite{Conventional_CycleGAN} to encourage the generators to identify inputs from specific domains, and approximate an identity mapping when target domain signals are used as inputs:
\begin{footnotesize}
\begin{equation}
\begin{aligned}
\label{identity}
L_{\text{idt}} \!=\!\mathbb{E}_{\text{x}_{\text{g}} \sim X_{\text{G}}}\!\!\left[\left\|G_{\text{N2G}}\left(x_{\text{g}}\right)\!-\!x_{\text{g}}\right\|_{1}\right] \!+ \!\mathbb{E}_{\text{x}_{\text{n}} \sim X_{\text{N}}}\!\!\left[\left\|G_{\text{G2N}}\left(x_{\text{n}}\right)\!-\!x_{\text{n}}\right\|_{1}\right]\!.
\end{aligned}
\end{equation}
\end{footnotesize}
Finally, the domain translation loss is:
\begin{footnotesize}
\begin{equation}
\begin{aligned}
\label{domain translation}
 & \nonumber L_{\text{tran}} =\lambda_{1} L_{\text{cyc}} + \lambda_{2} L_{\text{idt}}
\\ \nonumber &+L_{\text{GAN}}^{\text{img}}(X_{\text{N}},X_{\text{G}},D_{\text{N}}^{\text{img}},G_{\text{G2N}}) + L_{\text{GAN}}^{\text{feat}}(X_{\text{N}},X_{\text{G}},D_{\text{N}}^{\text{feat}},G_{\text{G2N}},F_{\text{N}}) \\
  &+ L_{\text{GAN}}^{\text{img}}(X_{\text{N}},X_{\text{G}},D_{\text{G}}^{\text{img}},G_{\text{N2G}})+  L_{\text{GAN}}^{\text{feat}}(X_{\text{N}},X_{\text{G}},D_{\text{G}}^{\text{feat}},G_{\text{N2G}},F_{\text{G}}) .
\end{aligned}
\end{equation}
\end{footnotesize}

\noindent
\textbf{Full Objective Function.}
The full objective function is defined as:
\begin{footnotesize}
\begin{equation}
\begin{aligned}
\label{total loss}
L &=L_{\text {\text{tran}}}+\lambda_{p}\left(L_{\text{pair}}\right)+\lambda_{c} L_{\text{blt}}.
\end{aligned}
\end{equation}
\end{footnotesize}
where $\lambda_{p}$ and $\lambda_{c}$ are trade-off weights that can be tuned to control the relative importance of the objectives.

\subsection{Progressive and Cooperative Learning}\label{train_Strategy}

\begin{algorithm}[t]
\footnotesize
    \captionsetup{font={footnotesize,it}}
    \caption{Progressive and Cooperative Training Strategy}
    \label{Algorithm_1}
    \begin{algorithmic}[0]
		\State \textbf{Step 1.} Train the bilateral domain translation module $G_{\text{G2N}}$ and $G_{\text{N2G}}$.
		\par
		\textbf{Input:} grayscale images $X_\text{G}$, NIR images $X_\text{N}$
		\par
		\textbf{Output:} latent spectrum translations $X_{\text{N2G}}=G_{\text{N2G}}(X_\text{N})$, $X_{\text{G2N}}=G_{\text{G2N}}(X_\text{G})$
		\par
		\textcolor{blue}{\textbf{Loss function:} $L_{\text{GAN}}^{\text{img}}$ (Eq.(\ref{GAN_pixel_N}), (\ref{GAN_pixel_G})), $L_{\text{cyc}}$ (Eq.(\ref{domain_cycle})), $L_{\text{idt}}$ (Eq.(\ref{identity}))}
		\par
		\vspace{1mm}
		
		\State \textbf{Step 2.} Train the grayscale colorization module $F_{\text{G}}$ and NIR colorization module $F_{\text{N}}$ parallelly.
		\par
		\textbf{Input:} $X_\text{G}$, $X_\text{N}$, $X_{\text{N2G}}$, $X_{\text{G2N}}$
		\par
		\textbf{Output:} Colorized images $F_\text{G}(X_\text{G})$, $F_\text{G}(X_{\text{N2G}})$, $F_\text{N}(X_\text{N})$, $F_\text{N}(X_{\text{G2N}})$
		\par
		\textcolor{blue}{\textbf{Loss function:} $L_{\text{pair}}$ (Eq.(\ref{pair loss})), $L_{\text{blt}}$ (Eq.(\ref{consistency loss})), $L_{\text{GAN}}^{\text{feat}}$ (Eq.(\ref{GAN_feature_N}), (\ref{GAN_feature_G}))}
		\vspace{1mm}
		
		\State \textbf{Step 3.} Fine-tune the pre-trained modules $G_{\text{G2N}}$, $G_{\text{N2G}}$, $F_{\text{G}}$ and $F_{\text{N}}$ jointly to formulate our whole model with a full loss function (Eq.(13)).
		\par
		\textbf{Input:} $X_\text{G}$, $X_\text{N}$
		\par
		\textbf{Output:} Colorized images $F_\text{G}(X_\text{G})$, $F_\text{G}(G_{\text{N2G}}(X_\text{N}))$, $F_\text{N}(X_\text{N})$, $F_\text{N}(G_{\text{G2N}}(X_\text{G}))$
		\par
		\textcolor{blue}{\textbf{Loss function:} $L$ (Eq.(\ref{total loss}))}
	\end{algorithmic} 
\end{algorithm}

We observe that when training the parameters of the four generators from scratch simultaneously, the network suffers from clear performance drops. The main reason is that, in the early stage, the features extracted by the bilateral domain translation module are of low quality and change significantly, making our two colorization modules unstably updated and failing to reach a local optimum.
To address this issue, inspired by \cite{kim2020learning} and \cite{pan2021exploiting}, we propose a progressive and cooperative training strategy that consists of the following three training phases:

\textbf{(i) Domain-translation training phase}. In the first training phase, we first independently explore the \textbf{latent spectrum context priors} by training the bilateral domain translation module using ground truth grayscale images $X_\text{G}$ and NIR images $X_\text{N}$ as supervision using the domain translation losses \cite{Conventional_CycleGAN} (Eq. (\ref{GAN_pixel_N}), (\ref{GAN_pixel_G}), (\ref{domain_cycle}), (\ref{identity})). This module generates latent spectrum translations $X_{\text{G2N}}$ and $X_{\text{N2G}}$.

\textbf{(ii) Parallel-colorization training phase}. In the second training phase, the \textbf{task domain priors} are explored by training the two colorization modules on both real images ($X_\text{N}$, $X_\text{G}$) and latent spectrum translations ($X_{\text{G2N}}$, $X_{\text{N2G}}$).
Note that the parallel-training stage executes two colorization tasks in parallel to learn from each other at pixel level with the supervised pair loss constraint $L_{\text{pair}}$ (Eq. (\ref{pair loss})) and also the perceptual level with the bilateral consistency constraint $L_{\text{blt}}$ (Eq. (\ref{consistency loss})). Meanwhile, two types of discriminators are adopted over the latent spectrum translations at the feature level. (Eq. (\ref{GAN_feature_N}), (\ref{GAN_feature_G})).
\par
\textbf{(iii) Cooperative fine-tune training phase}. In the last phase, all pre-trained modules are fine-tuned together cooperatively using the full objective function (Eq. \ref{total loss}), during which the aforementioned cross-domain priors are finally integrated. 
A detailed steps of our whole progressive and cooperative learning strategy are generalized in \textbf{Algorithm}~\ref{Algorithm_1}.


\begin{figure*}[t]
    \centering
  { 
      \includegraphics[width=0.975\linewidth]{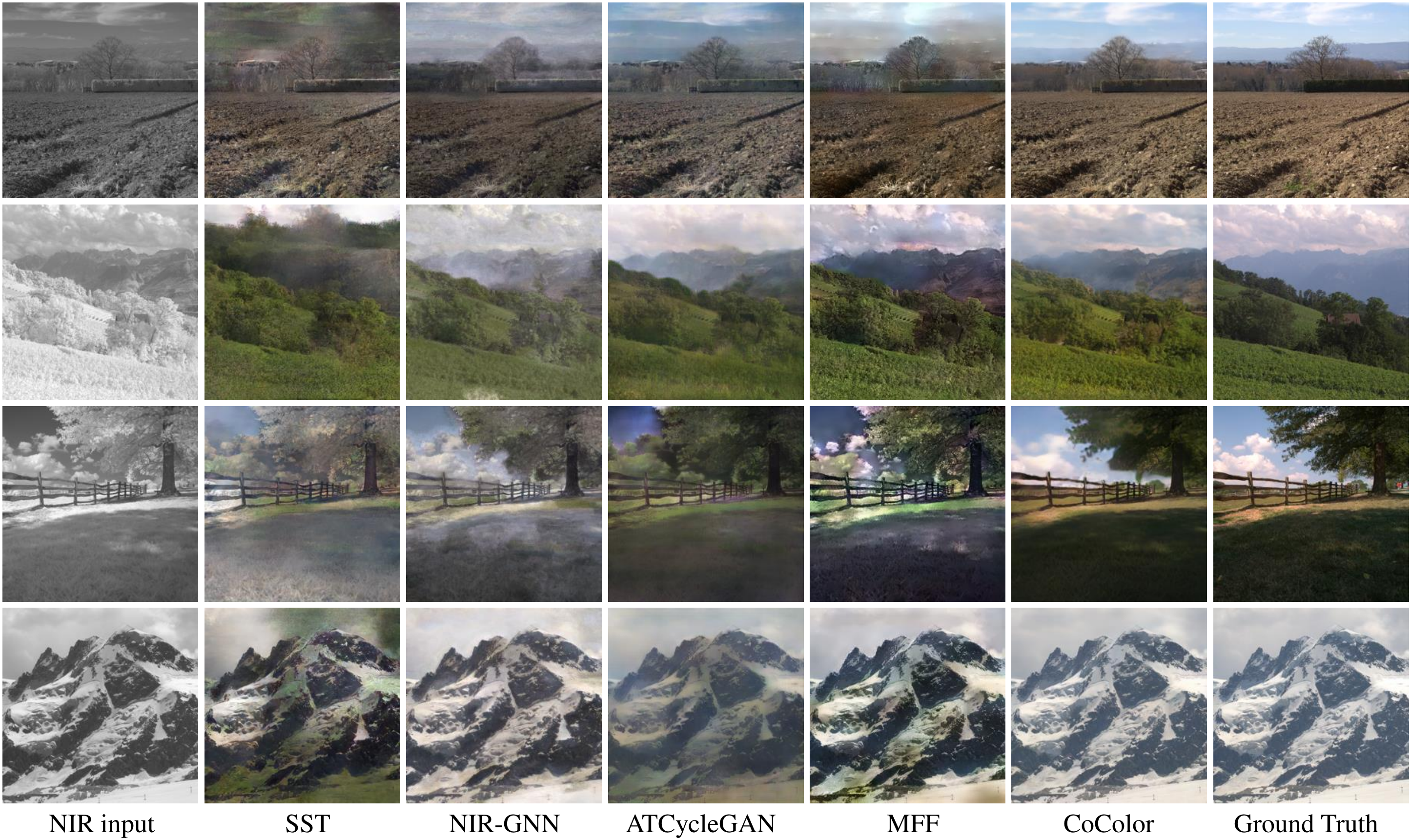}}
      \caption{Visual comparison among different methods on the VCIP testing dataset. Pictures from left to right correspond to NIR inputs, generated results of SST \cite{9301788}, NIR-GNN \cite{9301839}, ATCycleGAN \cite{9301791}, MFF \cite{9301787}, CoColor(our method) and ground truths.}
      \label{visual comparison} 
\end{figure*} 

\begin{figure*}[t]
    \centering
  { 
      \includegraphics[width=0.98\linewidth]{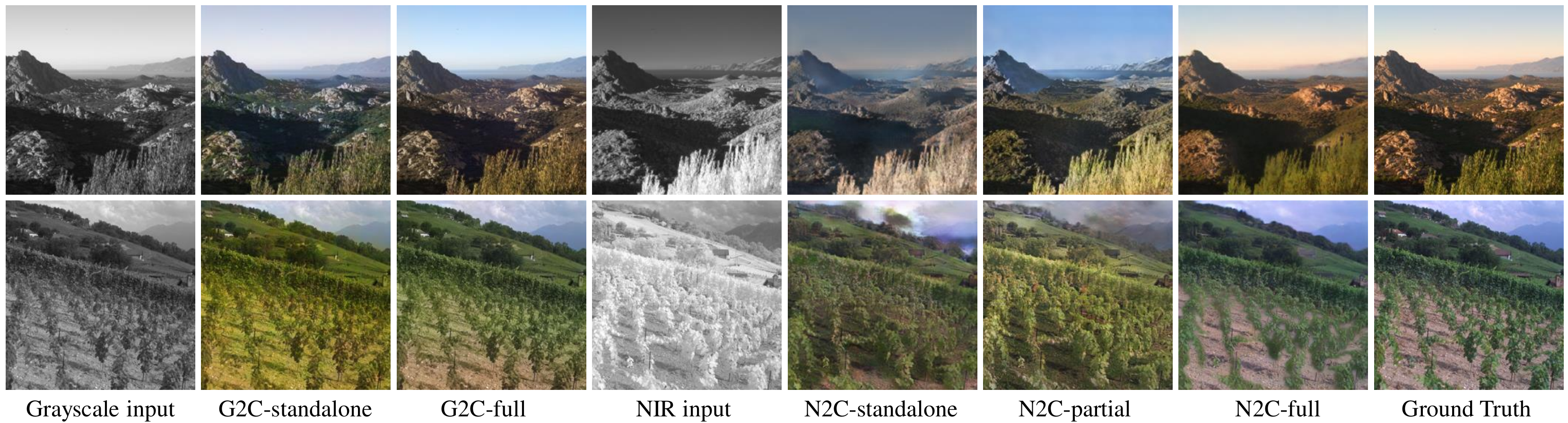}}
      \caption{Breakdown ablations of the full cooperative colorization-based framework with different sub-modules.}
      \label{ablation comparison} 
\end{figure*}

\section{Evaluation}
In this section, we first introduce the implementation details of our framework. Next, we evaluate both visual and quantitative results of our framework compared with the state-of-the-art methods of NIR and grayscale image colorization tasks based on two datasets. Finally, ablation studies are conducted to validate the proposed framework.

\noindent
\textbf{Implementation Details.}
We use the VCIP2020 Grand Challenge on the NIR image colorization dataset for both training and testing. Specifically, there are 372 NIR-RGB image pairs in the training dataset and another 28 pairs for testing.  
The ``RGB-Online'' subset is used for grayscale image colorization, which contains 1020 RGB images from the Internet in different landscape categories, including country, forest, field, and mountain. 
All the images have the same size of $256\times256$ pixels and are normalized to the range of $(0, 1)$. In order to make full use of the available data, we employ data augmentation by scaling, mirroring, random size cropping, and contrast adjustment.
\par
We implement our framework by Pytorch and utilize ADAM \cite{su2020instance} optimizer with a batch size of 10. 
As described in Section~\ref{train_Strategy}, we train the models in three phases. Firstly, we train the domain translation module for 400 epochs with a learning rate $l_{\text{tran}} = 1 \times 10^{-4}$. Next, we train our NIR image colorization network $F_\text{N}$ on $\left\{X_{\text{N}}, X_{\text{G2N}}\right\}$ and grayscale image colorization network $F_\text{G}$ on $\left\{X_{\text{G}}, X_{\text{N2G}}\right\}$ for 250 epochs with a learning rate  $l_{\text{c}} = 1 \times 10^{-4}$. At last, we fine-tune the whole network using the above pre-trained models. For the  hyperparameters in Eq.~(\ref{total loss}), we set $\lambda_{1} = 0.1$, $\lambda_{2} = 0.01$, $\lambda_{3} = 0.025$, $\lambda_{4} = 0.025$, $\lambda_{p} = 10$, and $\lambda_{c} = 1$, respectively.

\noindent
\subsection{Experiments on NIR Image Colorization}
\begin{table}[t]\small
\begin{center}
\caption{Quantitative comparison among different NIR image colorization methods. The best results are highlighted in \textbf{bold}.}
\centering
\label{result_table_1} 
\resizebox{1\columnwidth}{!}{
\begin{tabular}{lcccc}
\toprule
\multirow{2}{*}{Methods}   & \multicolumn{4}{c}{Metrics} \\
& PSNR ($\uparrow$) & SSIM ($\uparrow$) & AE ($\downarrow$) & LPIPS ($\downarrow$) \\ 
\toprule
\textit{NIR-GNN\cite{9301839}}~  & 17.50 & 0.60 & 5.22 & 0.384\\
\textit{MFF\cite{9301787}}~   & 17.39 & 0.61 & 4.69 & 0.318 \\
\textit{SST\cite{9301788}}~ & 14.26 & 0.57 & 5.61 & 0.361\\
\textit{ATCycleGAN\cite{9301791}}~ & 19.59 & 0.59 & 4.33 & 0.295 \\
\textit{\textbf{Ours}}  & \textbf{23.54}  & \textbf{0.69} & \textbf{2.68} & \textbf{0.223} \\
\bottomrule
\end{tabular}
}
\end{center}
\end{table}
We first evaluate our method comprehensively by comparing it with four state-of-the-art NIR image colorization approaches: ATCycleGAN \cite{9301791}, NIR-GNN \cite{9301839}, MFF \cite{9301787}, and SST \cite{9301788}. 
As shown in Fig. \ref{visual comparison}, all these methods have salient color distortion in all colorized images.
In contrast, benefiting from cooperative learning, as well as the progressive and cooperative training strategy, our method exploits the latent cross-domain priors between NIR and grayscale image colorization tasks, which expands the training dataset as well as guides the mapping process of \textit{NIR-to-RGB}. Thus, some ambiguous implicit mapping relationships for \textit{NIR-to-RGB} turn to be explicit.
For example, in the third row of Fig. \ref{visual comparison}, clouds and crowns of the tree in the NIR domain are difficult to be distinguishable, while they have different intensities in the grayscale domain. Our model, based on the proposed cooperative colorization, can colorize both clouds and crowns with the correct color while the other methods all fail. We also make quantitative comparisons illustrated in Table \ref{result_table_1}. Our method outperforms the other methods by a large margin. 

\begin{table}[t]\footnotesize
\begin{center}
\caption{Comparison with ATCycleGAN\cite{9301791} and DualGAN\cite{liang2021improved} over the EPFL dataset on the NIR image colorization task. The best results are highlighted in \textbf{bold}.}
\centering
\label{comparison_q_epfl} 
\resizebox{1\columnwidth}{!}{
\begin{tabular}{lcccc}
\toprule
\multirow{2}{*}{Methods}   & \multicolumn{3}{c}{Metrics} \\
& PSNR ($\uparrow$) & SSIM ($\uparrow$) & AE ($\downarrow$) & LPIPS ($\downarrow$) \\ 
\toprule
\textit{ATCycleGAN20'\cite{9301791}}~  & 16.89 & 0.53 & 5.64 & 0.382 \\
\textit{DualGAN21'\cite{liang2021improved}}~   & 17.80 & \textbf{0.62} & \textbackslash & \textbackslash \\
\textit{\textbf{Ours}}  & \textbf{20.14}  & 0.61 & \textbf{4.79} & \textbf{0.290} \\
\bottomrule
\end{tabular}
}
\end{center}
\end{table}

To further investigate the performance and generalization ability of our framework, we retrained both our network and ATCycleGAN \cite{9301791} using the EPFL dataset\cite{brown2011multi}, which has 477 NIR-RGB images pairs in total. Meanwhile, we also compared with the reported results of DualGAN\cite{liang2021improved}(training code is unavailable), which was also trained and evaluated on the same dataset. Note that the EPFL dataset has more complicated scene categories than the VCIP dataset, which involves urban, water, street, old buildings, and so forth. The quantitative results of the test set are shown in Table \ref{comparison_q_epfl}, respectively. Obviously, our method still outperforms these two methods by a large margin (Visual results are provided in the \textbf{\textit{Supplementary Material}}).

\begin{figure}[t]
    \centering
  { 
      \includegraphics[width=1\linewidth]{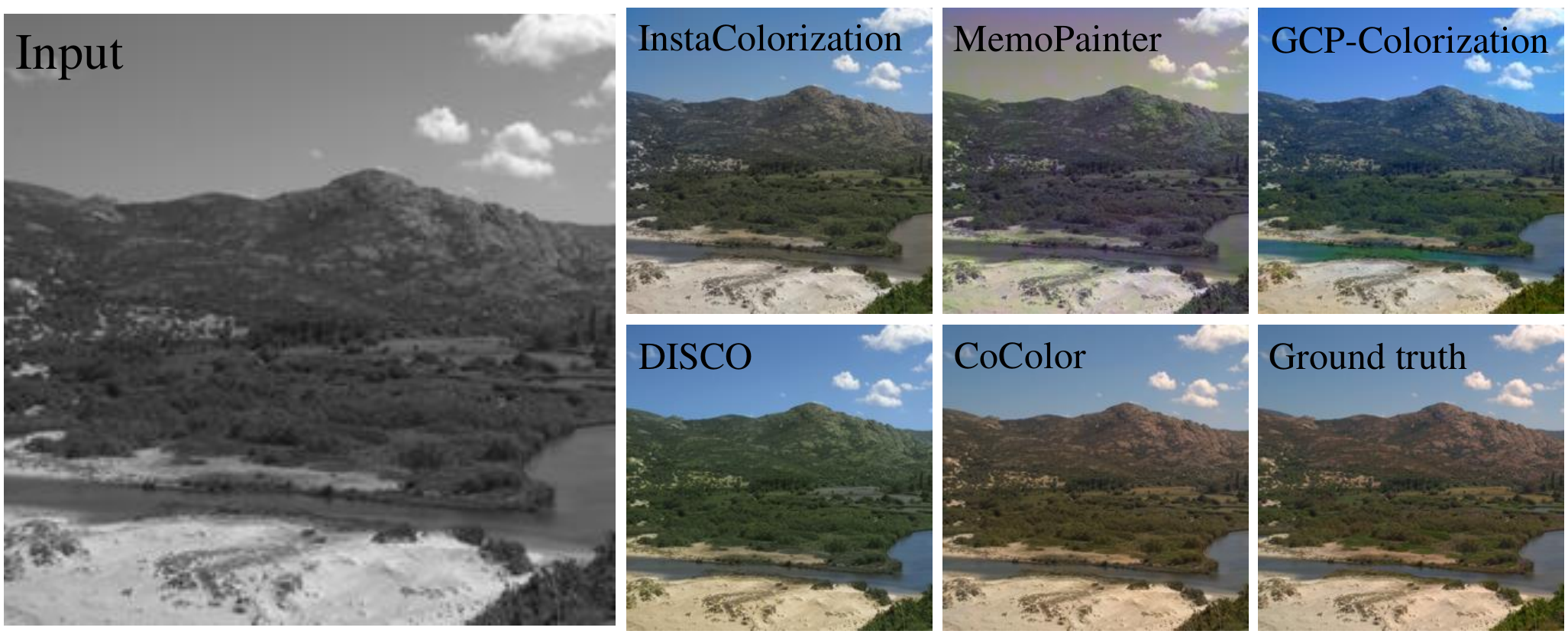}}
      \caption{Comparison with InstaColorization\cite{su2020instance}, MemoPainter\cite{yoo2019coloring}, GCP-Colorization\cite{wu2021towards} and DISCO\cite{xia2022disentangled} on the grayscale image colorization task.}
      \label{visual grayscale comparison} 
\end{figure}

\begin{figure}[t]
    \centering
  { 
       \includegraphics[width=1\linewidth]{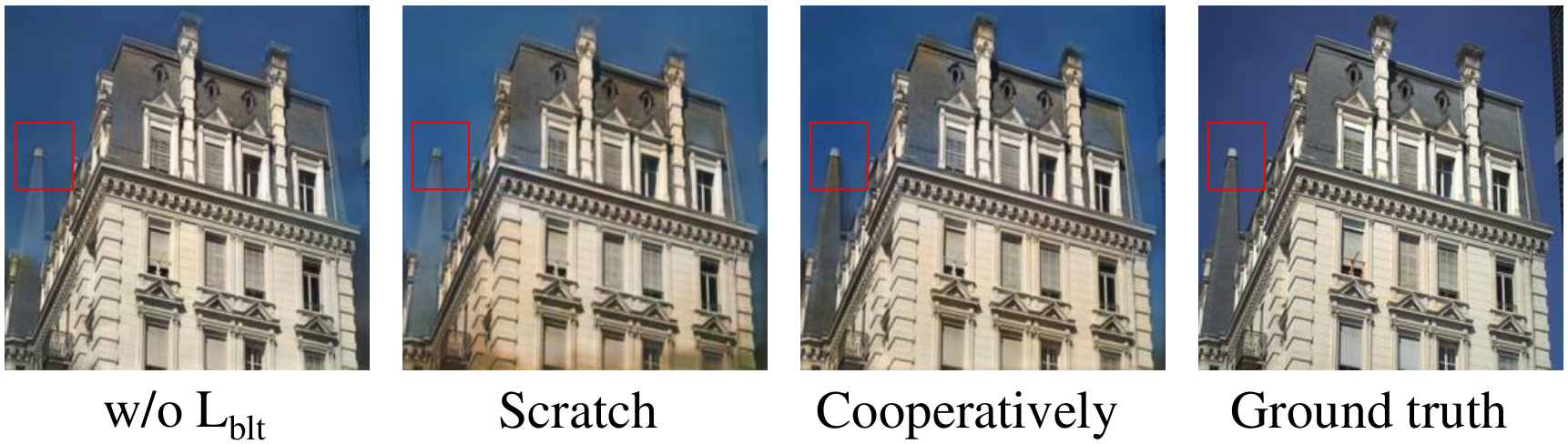}}
      \caption{Training strategies comparisons. The main differences are highlighted in the red box.}
      \label{ablation loss comparison} 
\end{figure} 

\begin{table}[t]\small
\begin{center}
\caption{Comparison with InstaColorization\cite{su2020instance}, MemoPainter\cite{yoo2019coloring}, GCP-Colorization\cite{wu2021towards} and DISCO\cite{xia2022disentangled} on the grayscale image colorization task. The best results are highlighted in \textbf{bold}.}
\centering
\label{comparison_gray} 
\resizebox{1\columnwidth}{!}{
\begin{tabular}{lcccc}
\toprule
\multirow{2}{*}{Methods}   & \multicolumn{3}{c}{Metrics} \\
& PSNR ($\uparrow$) & SSIM ($\uparrow$) & AE ($\downarrow$)  & LPIPS ($\downarrow$) \\ 
\toprule
\textit{InstaColorization20'\cite{su2020instance}}~  & 25.44 & 0.95 & 6.51 & 0.155 \\
\textit{MemoPainter19'\cite{yoo2019coloring}}~   & 28.56 & 0.93 & 5.39 & 0.138 \\
\textit{GCP-Colorization21'\cite{wu2021towards}}~  & 26.73 & 0.95 & 6.84 & 0.182 \\
\textit{DISCO22'\cite{xia2022disentangled}}~   & 27.31  & \textbf{0.96}  & 6.15 & 0.153 \\
\textit{\textbf{CoColor(Ours)}}  & \textbf{30.10}  & \textbf{0.96} & \textbf{4.63} & \textbf{0.105} \\
\bottomrule
\end{tabular}
}
\end{center}
\end{table}
\noindent
\subsection{Experiments on Grayscale Image Colorization}
To further verify the cooperative learning paradigm for the proxy task: Grayscale image colorization, we compare our method with InstaColorization \cite{su2020instance}, MemoPainter\cite{yoo2019coloring}, GCP-Colorization\cite{wu2021towards} and DISCO\cite{xia2022disentangled}. We retrained all these methods using the VCIP2020 grand challenge dataset and kept their default settings. 
As shown in Fig.~\ref{visual grayscale comparison}, InstaColorization\cite{su2020instance}, GCP-Colorization\cite{wu2021towards} and DISCO\cite{xia2022disentangled} portray the sky with blue color, as this is prior knowledge that can be easily learned by neural networks statistically, but it could be unrealistic and far from the corresponding RGB ground truths. Besides, MemoPainter\cite{su2020instance} fails to colorize the scene with vivid color.
In contrast, our cross-domain prior learning process of \textit{NIR-to-gray} introduces more diversity in the grayscale colorization phase, making it become difficult to learn such prior knowledge in a statistical sense, which compels our model to learn a more semantically reasonable mapping relationship. 
As a result, our model can generate more naturally colorized images. Quantitative results in Table \ref{comparison_gray} also demonstrate the superiority of our method.

\begin{table}[t]\normalsize
\begin{center}
\caption{Training strategy comparison on the EPFL dataset. The best results are highlighted in \textbf{bold}.}
\centering
\label{Ablation_table_2} 
\resizebox{1\columnwidth}{!}{
\begin{tabular}{lcccc}
\toprule
\multirow{2}{*}{Setting}   & \multicolumn{4}{c}{Metrics} \\
 & PSNR ($\uparrow$) & SSIM ($\uparrow$) & AE ($\downarrow$) & LPIPS ($\downarrow$)\\
\toprule
\textit{w/o $L_{\text{blt}}$}~  & 19.56 & \textbf{0.61} & 4.91 & 0.295\\
\textit{Training from scratch}~   & 19.47  & 0.60 & 4.95 & 0.319 \\
\textit{Training cooperatively}~   & \textbf{20.14} & \textbf{0.61} & \textbf{4.79} & \textbf{0.290} \\
\bottomrule
\end{tabular}}

\end{center}
\end{table}
\noindent
\subsection{Ablation Study}

\noindent
\textbf{1) Advantages of our cooperative training strategy.}
One of the ablation studies is to train the whole model from scratch(\textit{i.e.}, train the bilateral domain translation module and two colorization modules simultaneously).
In addition, to demonstrate the effectiveness of our cooperative training strategy, we investigate the role of our self-supervised bilateral consistency loss, by disabling the $L_{\text{blt}}$ (Eq.~\ref{consistency loss}) and keeping the other settings. The results of these two variations can be seen in Table \ref{Ablation_table_2} and Fig.~\ref{ablation loss comparison}.
As can be seen, the results produced by training w/o $L_{\text{blt}}$ and training from scratch sacrifice fine-grained contents and structural details. 
On the contrary, only the model trained cooperatively and within the bilateral consistency constraint can generate results with both vivid chrominance and correct semantics. 

Besides, to further validate the influence of different training strategies played on our bilateral domain translation module, we provide the visual results generated by three different strategies: pre-training of the bilateral domain translation module, training the whole model from scratch, and training the whole model using our progressive and cooperative learning strategy, as shown in Fig. \ref{evaluation_DA}. As can be seen, our cooperative learning strategy further improves the performance of the bilateral domain translation module(compare Fig. \ref{evaluation_DA}(b) and Fig. \ref{evaluation_DA}(d)).

\noindent
\textbf{2) Effectiveness of each module.}
We designed several pipelines to verify the effectiveness of each module in our framework, including $1)$ \textbf{N2C-\textit{full}}: the full framework applied on NIR image colorization; $2)$ \textbf{G2C-\textit{full}}: the full framework applied on grayscale image colorization; $3)$ \textbf{N2C-\textit{partial}}: the NIR image colorization network using N2C and N2G2C; $4)$ \textbf{N2C-\textit{standalone}}: the NIR image colorization network using N2C; $5)$ \textbf{G2C-\textit{standalone}}: the grayscale image colorization network using G2C.

Qualitative and quantitative results are illustrated in Fig.~\ref{ablation comparison} and Table \ref{ablation_table_1}, respectively. 
As shown in the fifth column in Fig.~\ref{ablation comparison}, the \textbf{N2C-\textit{standalone}} network suffers from color misalignment and distortion due to mapping ambiguity and data insufficiency. In contrast, \textbf{N2C-\textit{partial}} network produces results with better color consistency benefiting from the latent spectrum translation generated by grayscale images via domain translation, and thus, the insufficient data problem can be relieved. However, it still fails in learning the mapping  between the NIR domain and the RGB domain in some more complicated scenarios from the semantic level (\textit{e.g.}, clouds in the second row are misaligned). Overall, only using our full cooperative learning framework, both mapping ambiguity and data insufficiency problems can be well solved.


\begin{table}[t]\small
\begin{center}
\caption{Breakdown ablations. The best results of each task are highlighted in \textbf{bold}.}
\centering
\label{ablation_table_1} 
\resizebox{1\columnwidth}{!}{
\begin{tabular}{lcccc}
\toprule
\multirow{2}{*}{Pipeline}   & \multicolumn{4}{c}{Metrics} \\
& PSNR ($\uparrow$) & SSIM ($\uparrow$) & AE ($\downarrow$) & LPIPS ($\downarrow$) \\ 
\toprule
\textit{G2C-standalone}~  & 28.68 & 0.94 & 5.43 & 0.171\\
\textit{G2C-full}~   & \textbf{30.10} & \textbf{0.96} & \textbf{4.63} & \textbf{0.105} \\
\textit{N2C-standalone}~   & 18.88 & 0.59 & 5.81 & 0.305 \\
\textit{N2C-partial}~   & 20.24 & 0.63 & 5.49 & 0.282\\
\textit{N2C-full}~   & \textbf{23.54} & \textbf{0.69} & \textbf{2.68} & \textbf{0.223} \\
\bottomrule
\end{tabular}
}
\end{center}
\end{table}

\subsection{Limitations}
\begin{figure}[t]
    \centering
  { 
      \includegraphics[width=1\linewidth]{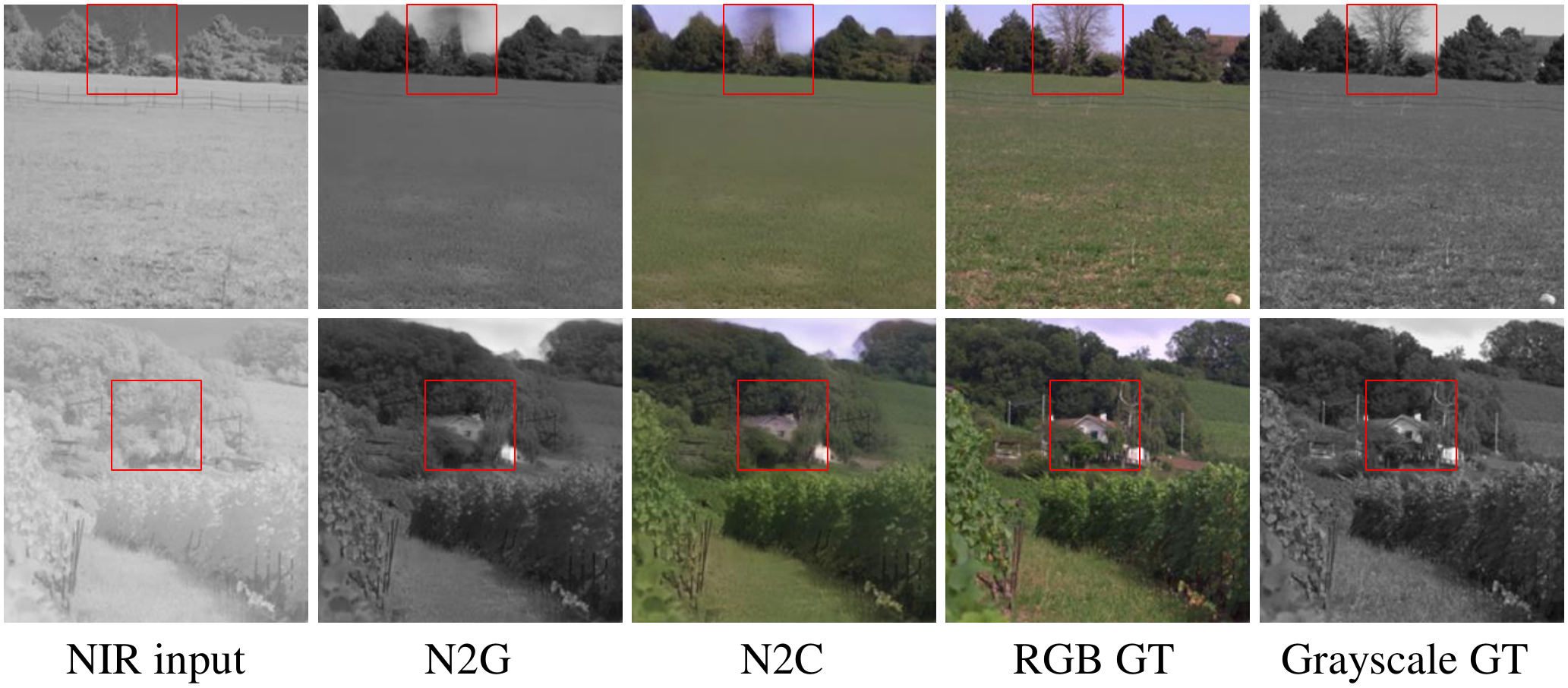}}
      \caption{Failure cases generated by our CoColor framework. The main difference is highlighted in the red box.}
      \label{failure_case} 
\end{figure}

Colorizing NIR and grayscale images while preserving both semantic and texture details is a challenging task, particularly when the intensity and texture differences between the NIR and grayscale domains are significant. Despite our best efforts, some unsatisfactory results were generated by our framework, as shown in the examples in Fig.  \ref{failure_case}. For instance, in the second row of the figure, the intensity of the house in the NIR domain is similar to that of the background trees, and lacks clear texture details. Consequently, the grayscale intermediate latent spectrum translation is blurry and fails to serve as an effective cross-domain prior for NIR-to-RGB spectrum translation. As a result, the structure and color of the house are not generated accurately.


\section{Conclusions}
In this paper, we propose a novel cooperative learning framework for efficient NIR image spectrum translation, which contains a bilateral domain translation module between the NIR domain and the grayscale domain, as well as two colorization modules to colorize grayscale images and NIR images, respectively, to explore the latent cross-domain priors for NIR image colorization. 
Benefiting from the progressive and cooperative learning strategy, the NIR training set can be expanded by numerous latent spectrum translations from grayscale images, which act as the latent spectrum context priors; Meanwhile, the mapping ambiguity can be relieved by learning from different task domain priors. Extensive experimental results on both NIR image colorization and grayscale image colorization demonstrate that our algorithm outperforms the state-of-the-art approaches. 
We hope this work can provide some insights for general spectrum reconstruction challenges (\textit{e.g.}, hyperspectral reconstruction from RGB images) and small-scale sample learning tasks (\textit{e.g.}, aligning domain shift between synthetic data and real data in transfer learning).


\bibliographystyle{ACM-Reference-Format}
\balance
\bibliography{acmart}

\end{document}